\documentclass[letterpaper, 10 pt, conference]{ieeeconf}
\IEEEoverridecommandlockouts
\usepackage{cite}
\usepackage{amsmath,amssymb,amsfonts}
\usepackage{algorithmic}
\usepackage{graphicx}
\usepackage{textcomp}
\usepackage{xcolor}

\usepackage{ctable}
\usepackage{booktabs}
\usepackage{multirow}
\usepackage{ctable}
\usepackage{subcaption}
\usepackage{comment}

\usepackage{bibspacing}
\def\BibTeX{{\rm B\kern-.05em{\sc i\kern-.025em b}\kern-.08em
    T\kern-.1667em\lower.7ex\hbox{E}\kern-.125emX}}

\title{A Deep Learning Approach to Predict Blood Pressure from PPG Signals}
\author{Ali Tazarv$^{\star}$, Marco Levorato$^{\dagger}$\\
${\star}$ EECS Department, ${\dagger}$ Dept. of Computer Science\\
University of California, Irvine
\vspace{-3ex}}
\begin{document}

\maketitle

\thispagestyle{empty}
\pagestyle{empty}

\begin{abstract}
Blood Pressure (BP) is one of the four primary vital signs indicating the status of the body's vital (life-sustaining) functions. BP is difficult to continuously monitor using a sphygmomanometer (i.e. a blood pressure cuff), especially in everyday-setting. However, other health signals which can be easily and continuously acquired, such as photoplethysmography (PPG), show some similarities with the Aortic Pressure waveform. Based on these similarities, in recent years several methods were proposed to predict BP from the PPG signal. Building on these results, we propose an advanced personalized data-driven approach that uses a three-layer deep neural network to estimate BP based on PPG signals. Different from previous work, the proposed model analyzes the PPG signal in time-domain and automatically extracts the most critical features for this specific application, then uses a variation of recurrent neural networks called Long-Short-Term-Memory (LSTM) to map the extracted features to the BP value associated with that time window. Experimental results on two separate standard hospital datasets, yielded absolute errors mean and absolute error standard deviation for systolic and diastolic BP values outperforming prior works.
\end{abstract}

% \begin{IEEEkeywords}
% PPG, Blood Pressure, Deep Learning, Feature Extraction, Convolutional Neural Network, Long-Short-Term-Memory.
% \end{IEEEkeywords}

%%%%%%%%%%%%%%%%%%%%%%%%%%%%%%%%%%%%%%%%%%%%%%%%%%%%%%%%%%%%%%%%%%%%%%%%%%%%%%%%
\section{Introduction}
Hypertension, defined as systolic blood pressure (SBP) larger than 140mmHg or diastolic blood pressure (DBP) larger than 90mmHg\footnote{for a patient who is not undergoing drug treatment for hypertension.} 
\cite{doi:10.1161/CIR.0000000000000558}, is estimated to have caused $9.4$ million annual deaths globally, $17$\% of the total death in 2012 and $7$\% of total disability-adjusted life years (DALYs) \cite{world2014global}. If left uncontrolled,
hypertension causes stroke, myocardial infarction (MI), cardiac failure, dementia, renal failure, and even blindness. In adults, hypertension after diabetes is the second reason to increase the risk of cardiovascular disease (CVD) and several types of cancer, as well as multiple non-fatal diseases. Hypertension has been increasing in recent years. By 2030, $40.5$\% of the US population is projected to have some form of CVD \cite{doi:10.1161/CIR.0b013e31820a55f5}.
%The prevalence of hypertension is highest in some low-income countries \cite{world2014global}. 
While people with the risk of hypertension need to measure their blood pressure frequently, conventional cuff-based BP measurement devices are expensive and inconvenient for continuous monitoring. Thus the development of alternative methods is necessary.

\begin{figure}[b]
    \centering
    \vspace{-4.5ex}
    \includegraphics[width=0.4\textwidth]{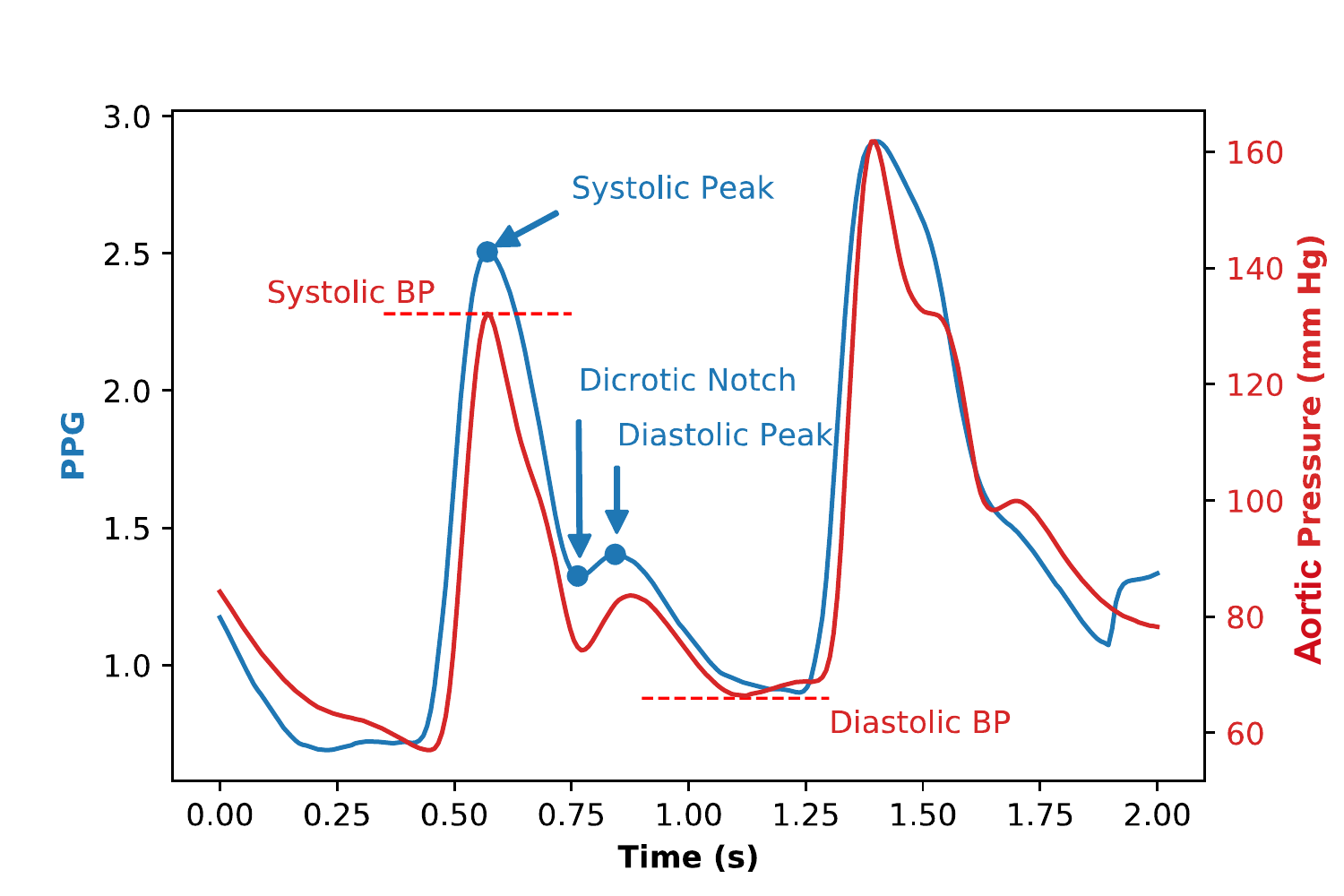}
    \caption{Aortic Pressure and PPG raw signals}
    \vspace{-1ex}
    \label{fig: PPG_BP_Cor}
\end{figure}

Blood Pressure (BP) -- commonly measured in mmHg -- is a quasi-periodic signal in sync with an individual's heartbeats. The upper peak in each period is called the Systolic Blood Pressure or SBP, and the lower bound in each period is called Diastolic Blood Pressure or DBP (Figure \ref{fig: PPG_BP_Cor}). While blood pressure is difficult to monitor continuously in a non-clinical setting, Photoplethysmography (PPG) is a non-invasive optical method that measures a related signal: blood volume temporal variations in the vessels and tissues.
PPG signals are obtained from pulse oximeters, emitting visible light (LED) on the skin and measuring the micro-variations in the transmitted,
or reflected light intensity (photo-diode). PPG sensors are small in size and low cost to build, and they already exist in most newer wearables (e.g. smartwatches, activity trackers, and smart rings). 

In recent years, there has been an extensive body of research studying similarities and correlation of PPG signals and Aortic Pressure waveforms, as well as the possibility of estimating SBP and DBP based on PPG signals.
Since both signals are originating from the same source (the individual's heartbeats), they are highly correlated. However, since Aortic and PPG are generally measured from different parts of the body (e.g. arm and wrist) using different devices, they are typically out of phase. Figure \ref{fig: PPG_BP_Cor} is based on experimental data after time-shift alignment (since PPG signal does not have a unit, in the figure it is scaled for easier readability). Some methods such as the one in~\cite{xing2016optical} are proposed to automatically detect and compensate for this phase difference. 
%The following has two figures (squeezed in the form of one). I kept one of them only:
% \begin{figure}[ht]
%      \centering
%      \begin{subfigure}[b]{.45\textwidth}
%          \centering
%          \includegraphics[width =1\textwidth]{Figures/2_PPG_BP.pdf}
%          \caption{Signals in the time Domain}
%          \label{fig: PPG_BP_Cor1}
%      \end{subfigure}
%      \begin{subfigure}[b]{0.45\textwidth}
%          \centering
%          \includegraphics[width =1\textwidth]{Figures/3_PPG_BP_Cor.pdf}
%          \caption{(PPG, ABP) samples}
%          \label{fig: PPG_BP_Cor2}
%      \end{subfigure}
%         \caption{ABP and PPG Correlation}
%         \label{fig: PPG_BP_Cor}
% \end{figure}
%
% \begin{figure}[b]
%     \centering
%     \includegraphics[width=.45\textwidth]{Figures/PPG-sensor.PNG}
%     \caption{\small Schematics of PPG instrument design}
%     \label{fig: ppg-sensor}
% \end{figure}

Due to convenience of use and also low cost of PPG sensors, estimating blood pressure metrics (SBP and DBP) from PPG signals is of great interest. The common approach for this is to first extract a set of predefined features from PPG, and then use some regression models to estimate BP from those features. A shortcoming of this approach is that since these features are predefined and generic, some of the important information inside the raw PPG signals (including patterns specific to each subject) might be lost, which if captured, could improve the accuracy of BP estimation. Other than that, these methods do not utilize characterizations of PPG as a time-series signal.
In this paper, we propose the use of a new framework to address these shortcomings. 

Our proposed approach makes the following advancements with respect to prior works in the literature:
(\emph{i}) The proposed method utilizes a Convolutional Neural Network (CNN) layer, which ``learns'' the most informative features of the PPG signal for this application (BP estimation) in a nonlinear and efficient way, as part of an optimization problem; (\emph{ii}) We adopt a Long-Short Term Memory (LSTM) model to capture long-term temporal inter-dependencies in a seamless and automated procedure.

We train and test the model for each subject separate from others. We test the proposed framework on two standard hospital datasets. Our proposed model on 20 randomly picked subjects from MIMIC-II dataset \cite{doi:10.1161/01.CIR.101.23.e215} gives prediction error ($MAE \pm SDAE$) of $3.70\pm3.07$ mmHg and $2.02\pm1.76$ mmHg, for SBP and DBP values respectively. 
Also on UQVSD dataset, prediction errors ($MAE \pm SDAE$) are $3.70\pm3.07$ mmHg and $2.02\pm1.76$ mmHg, for SBP and DBP respectively.
To the best of our knowledge, these results on similar datasets outperform other methods in the literature.

%which "learns" the most important features of the PPG signal in an efficient and smart way, for every subject separately, then uses an RNN to predict BP based on temporal behaviour of the extracted features. \\
The rest of the paper is organized as follows. Section~\ref{sec: background} overviews recent work in this area. Section~\ref{sec: Methodology} describes our proposed framework and methodology. %be consistent to 'estimation' or 'prediction' and use it in the introduction.
Section~\ref{sec: Data-set} introduces the two real-world datasets we used. In Section \ref{sec: results} we present the results and compare those with state-of-the-art methods. Section~\ref{sec: Conclusions} concludes the paper.

\section{Related Work} 
\label{sec: background}
%Considering the setup of PPG sensors, they are vulnerable to motion artifacts (MA) which result in distortion in the signal fidelity.
%MA originate from a number of factors \cite{elgendi2012analysis}; the physical activity which might change the location of the sensor on the skin, light leaking through the gap between the sensor and the skin, and the change in the blood flow because of the physical movements are the most important factors of MA. The overall effect is that MA might dominate the original PPG signal, which was intended to be an indicator of the heartbeats. Properties of PPG signal and some of its applications are reviewed in  \cite{allen2007photoplethysmography}.
Estimation of Blood Pressure from PPG signal is studied in prior works and several methods are proposed for that. In \cite{7168806}, \cite{lass2004continuous} and \cite{Miao2017}, the authors propose methods based on Pulse Transit Time (PTT). PTT is defined as the time required for the blood pressure wave to travel from the source (heart) to the wrist (where the PPG signal is recorded, and it is measured from the time shift between the Electrocardiography (ECG) and PPG signals. In \cite{MOUSAVI2019196}, authors consider every sample point in one cycle as a separate feature and use Principal Component Analysis (PCA) to reduce the number of features, and then apply regression algorithms on the set of reduced features. In~\cite{shimazaki2018features}, authors propose a list of some potential features of the PPG signals (e.g. area under the curve, time length of certain points on the signal in one cycle, etc.), and then use an autoencoder to reduce the list of features and at the end estimate BP based on the reduced set of features using a feed-forward neural network. 

All these methods start by extracting a set of features from PPG signals and then use those features to estimate blood pressure. 
One shortcoming of this approach is that since these features are predefined, they may fail to capture all the details in the signal that might be useful for a specific task such as BP estimation. Some of these details might also be different from one subject to another.
Even in \cite{MOUSAVI2019196} that the authors start with the whole signal, the dimensionality reduction procedure is not optimized for the BP estimation task, as PCA is a generic data compression method.
Additionally, none of these approaches exploit possible characterizations of the PPG as a time-series signal.
% In addition to BP estimation, other popular applications of PPG signals are Heart Rate monitoring \cite{zhang2014troika}\cite{8607019}, arterial oxygen saturation \cite{allen2007photoplethysmography}, Bio-metric Identification \cite{8607019}, determining arterial stiffness \cite{huotari2011photoplethysmography} and other bio-medical applications \cite{s18061894}.

%%%%%%%%%%%%%%%%%%%%%%%%%%%%%%%%%%%%%%%%%%%%%%%%%%%%%%%%%%%%%%%%%%%%%%%%%%%%%%%%
\section{Methodology} 
\label{sec: Methodology}
\begin{figure}[bt]
    \centering
    \includegraphics[width=0.48\textwidth]{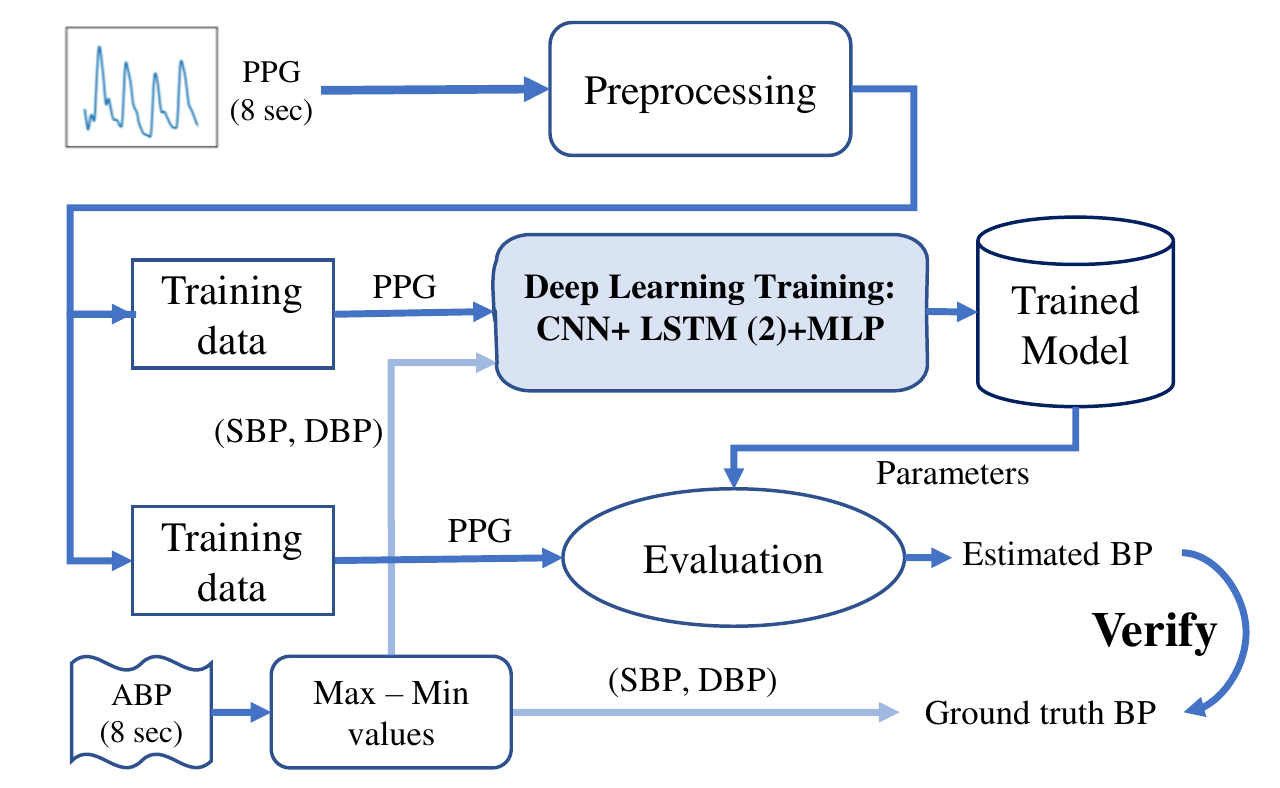}
    %\vspace{-4ex}
    \caption{Overview of the proposed method to estimate BP from single channel PPG signal}
    \label{fig: blockdiagram}
    \vspace{-2ex}
\end{figure}
Different from the common approach described in Section \ref{sec: background}, our proposed method for each subject automatically extracts a set of features that are optimal for this application, as part of the training process. It also captures and utilizes temporal dependencies in the data using an advanced neural architecture.
Figure \ref{fig: blockdiagram} presents the block diagram of the method we propose to estimate SBP and DBP using PPG signal. The method is composed of the following blocks and components:
(\emph{A}) Signal Pre-processing; (\emph{B}) Machine Learning, composed of an automatic feature extraction module (CNN) and a time series analysis module (LSTM); and (\emph{C}) Model Evaluation block. We explain the first two components here and describe the Evaluation block later in Section \ref{sec: results}.

\subsection{Pre-processing}
Since both Aortic Pressure and PPG originate from heartbeats, we do not expect to see high-frequency elements in their waveforms. Thus, higher frequency elements are likely due to noise. Additionally, since the PPG is a relative signal and its DC offset (mean value) does not have an interpretation of our interest, we eliminate the mean value of the signal (the zero frequency element). There are many popular methods to eliminate a range of frequencies from a signal. Herein, we use a traditional Fourier Transform (FFT)-based approach. 
%In the very beginning, we eliminate the high-frequency elements of the BP signal using a 4th order Butterworth lowpass filter with cut-off frequency 5 Hz. After that we split the PPG and the BP signals in 8 s windows (frame), sliding by 2 s (so 6s overlap). 
In the PPG signal, we set a band-pass filter eliminating the frequencies outside the range $0.1-8$~Hz. The resulting waveform still contains all the information necessary for our estimation, while being affected by smaller noise energy. In preprocessing the Aortic Pressure signal, we apply a low-pass filter with a cut-off frequency of $5$Hz to the raw signal to eliminate sharp peaks in the signal. After these steps, we split both the PPG and ABP into windows of $8$~s with a step of $2$s, resulting in $6$s overlaps between two consecutive windows.
%(gray shaded part in Figure \ref{fig: Overlap}). 
% \begin{figure}[bp]
%     \centering
%     \vspace{-3ex}
%     \includegraphics[width=.45\textwidth]{Figures/Overlap.pdf}
%     \caption{Splitting PPG signals into 8s windows, \\ \centering $t_i$ is the sample at timestamp $i$.}
%     \label{fig: Overlap}
% \end{figure}
We further scale the PPG signals in each window to zero mean and unit variance.
Each window of the PPG signal is equivalent to a vector of length $8f_s$, which $f_s$ is the sampling frequency. We use this vector as the model input.
In the ABP signal, the maximum and minimum values in each window are interpreted as the SBP and DBP respectively for that interval (Figure \ref{fig: PPG_BP_Cor}).
%and the resulting signal is passed through a 1 Hz digital lowpass filter (to eliminate large instant changes in the BP), and the result is considered as Systolic Blood Pressure (SBP) and Diastolic Blood Pressure (DBP) respectively.
\subsection{Deep Learning}
Deep Neural Networks primarily including Convolutional Neural Network (CNN), Recurrent Neural Network, and Multi-Layer-Perceptron (MLP), facilitate task-adapted feature representation of the data. CNN, which was originally designed to capture the variability of 2D data, in~\cite{726791} is shown to perform well on 1D data and outperform other approaches. CNN consists of an initial layer of convolutional filters -- boxes with certain sizes sF, and certain sets of trainable weights which slide over the input data -- followed by an activation function (e.g. sigmoid), a pooling layer with pool size sP, and after that a batch normalization layer \cite{DBLP:journals/corr/IoffeS15}. 
%Stacking multiple convolutional layers results in a non-linear, task-adaptive and efficient method to extract features from data \cite{10.1007/978-3-642-21735-7_7}.

RNN is an effective tool for the analysis of time-series data since it learns contextual patterns in the input from previous time steps. Unlike feed-forward networks, RNN uses some internal states to process temporal sequences of the input. However, learning to store information over extended time intervals via recurrent back-propagation may necessitate an excessive training time, due to the insufficient decaying error gradient back-flow (vanishing gradient) discussed in \cite{doi:10.1162/neco.1997.9.8.1735}.

\textit{Long-Short-Term-Memory} (LSTM) is a variation of RNN and solves this issue by using memory blocks, for which the trainable ``forget gate'', ``input gate'' and ``output gate'' control which parts of the data are worth saving and which parts are not. This idea solves the issue of vanishing gradient mentioned above, and the result is a strong tool to learn and track long-term inter-dependencies in the input data (e.g. time series data) \cite{doi:10.1162/neco.1997.9.8.1735}.
The combination of CNN with a subsequent LSTM has been shown to perform well on PPG signals. In \cite{8607019} authors use a similar architecture for estimating heart-rate and Bio-metric ID using PPG.
%Herein, we use a similar structure applied to the problem at hand.

The model is trained in Python environment (version 3.6.3) and implemented using Keras 2.2.4 \cite{chollet2015keras}, with Tensorflow 1.3.1 backend. The CNN layer starts with a 1-D filtering operation with filter size $15$ (sF), followed by a Rectified Linear Unit (RELU) \cite{Nair:2010:RLU:3104322.3104425} as activation function, a batch-normalization layer \cite{DBLP:journals/corr/IoffeS15}, a max-pooling layer, and finally a dropout layer \cite{JMLR:v15:srivastava14a} with $dropout\_rate = 0.1$. In the max-pooling layer the pooling size $sP$ is set to 4, which means the layer takes the maximum out of every 4 consecutive values in a filter (non-overlapping) and discards the rest. Unlike some other common deep network layers (e.g. MLP), the Max-pooling layer does not have trainable weights. Dropout layers are probabilistic masks that in every gradient update (mini-batch) block a portion of the nodes. In our implementation, the $dropout\_rate =0.1$ means that in every gradient update, there is a $10\%$ chance for each node in the layer to be dropped out (or masked).
These two-layers (Max-pooling and Dropout) are well-known methods used to avoid overfitting.

Our LSTM network is composed of two identical LSTM module in series. Each of them has $64$ units (nU), with \textit{tanh} as the activation function for the hidden state data and output data, and \textit{hard-sigmoid} as the recurrent activation functions for the forget, input, output gates \cite{doi:10.1162/neco.1997.9.8.1735}.

The model was trained using Adam optimizer \cite{Kingma2014AdamAM}. The batch size was set to $20$ to balance the training time vs noisiness of gradient updates trade-off. The hyperparameters mentioned above and also the number of layers for each network (CNN and LSTM), were optimized with grid search. Changing the hyperparameters (e.g. sF, nU, sP, etc.) or the number of layers did not result in an improved final output.
%%%%%%%%%%%%%%%%%%%%%%%%%%%%%%%%%%%%%%%%%%%%%%%%%%%%%%%%%%%%%%%%%%%%%%%%%%%%%%%%

\section{Dataset} \label{sec: Data-set}
To evaluate the accuracy and efficiency of the proposed method, we used two publicly accessible real-life datasets.

\begin{figure}[bt]
    \centering
    \includegraphics[width=0.9\linewidth]{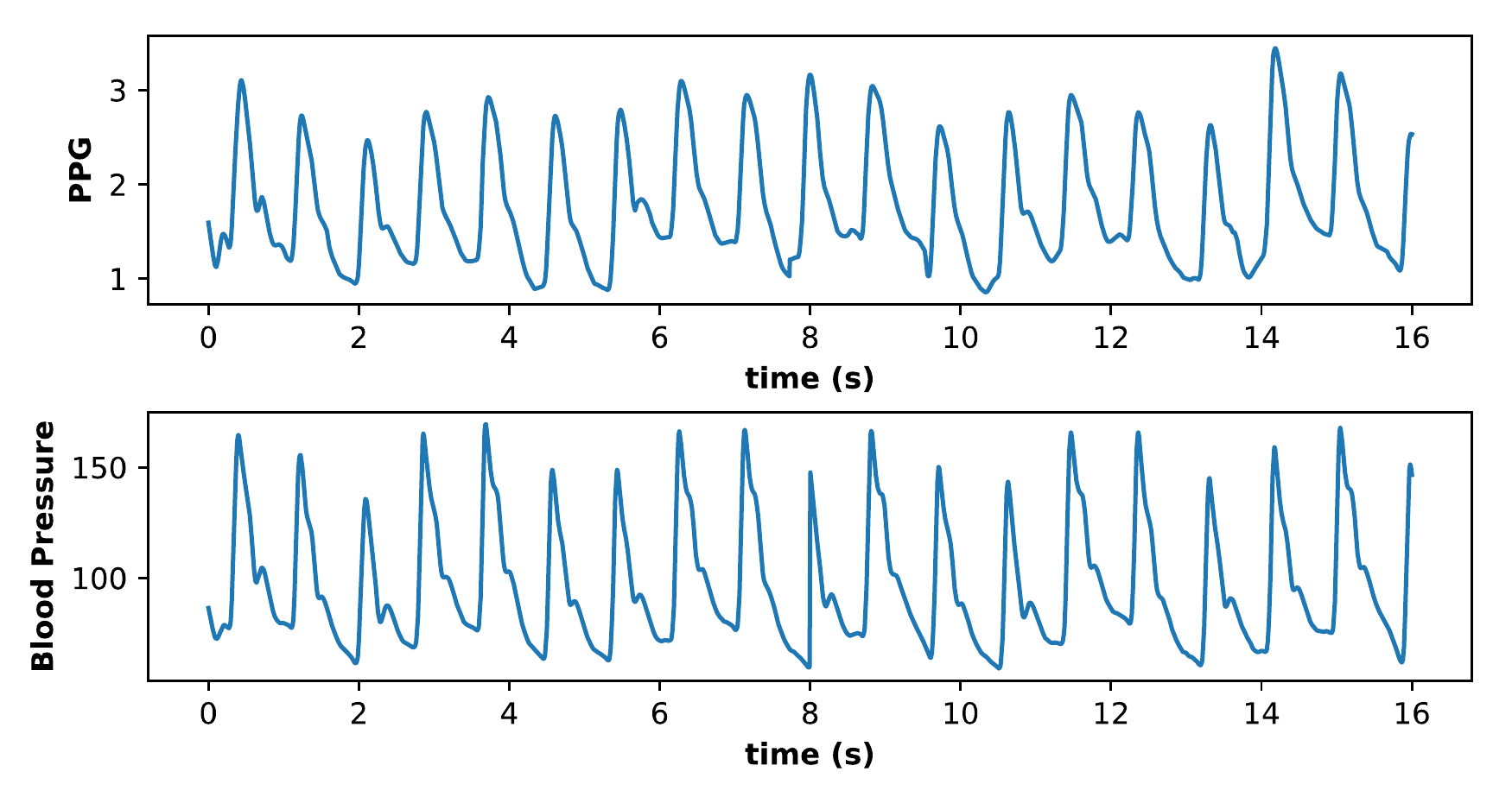}
    \caption{Scaled raw PPG and ABP signals in the same time interval, from MIMIC-II dataset.}
    \label{fig:PPG_BP_t}
    \vspace{-4ex}
\end{figure}

\textbf{MIMIC V.3 2015:} Multi-parameter Intelligent Monitoring in Intensive Care, provided by \cite{doi:10.1161/01.CIR.101.23.e215}. The MIMIC-II dataset includes healthcare information and signals of thousands of patients at hospitals between the years 2001 and 2012. 
%Part of this large dataset, including PPG, Arterial Blood Pressure (ABP), and Electrocardiogram (ECG) signals is extracted and used in another work \cite{7168806}. 
An example of PPG and corresponding ABP signals from this dataset is given in Figure \ref{fig:PPG_BP_t} (the PPG signal is scaled for readability). 
These signals were originally recorded at a sampling frequency $f_s$=125~Hz, with a minimum accuracy of 8 bits. We randomly selected data corresponding to 20 different subjects -five minutes long each- from this dataset to test our model.

\textbf{The University of Queensland Vital Signs Dataset:} This dataset covers a wide range of BP values, recorded from 32 surgical patients in duration ranging from 13 minutes to 5 hours, at the Royal Adelaide Hospital, Australia \cite{liu2012university}. This dataset is recorded at the sampling frequency $f_s=100 Hz$. We used 49 measurements (10 minutes each) from this dataset.

The majority of studies on this topic have used parts of one of these two datasets. However, details of how portions of these datasets were selected for experiments are not revealed. We acknowledge that the data we used for our experiment, might not exactly match the data used in similar studies, but in order to conduct a fair comparison, we picked a portion of each dataset randomly.

\begin{figure}[bp]
    \centering
    \vspace{-2ex}
    \includegraphics[width =.48\textwidth]{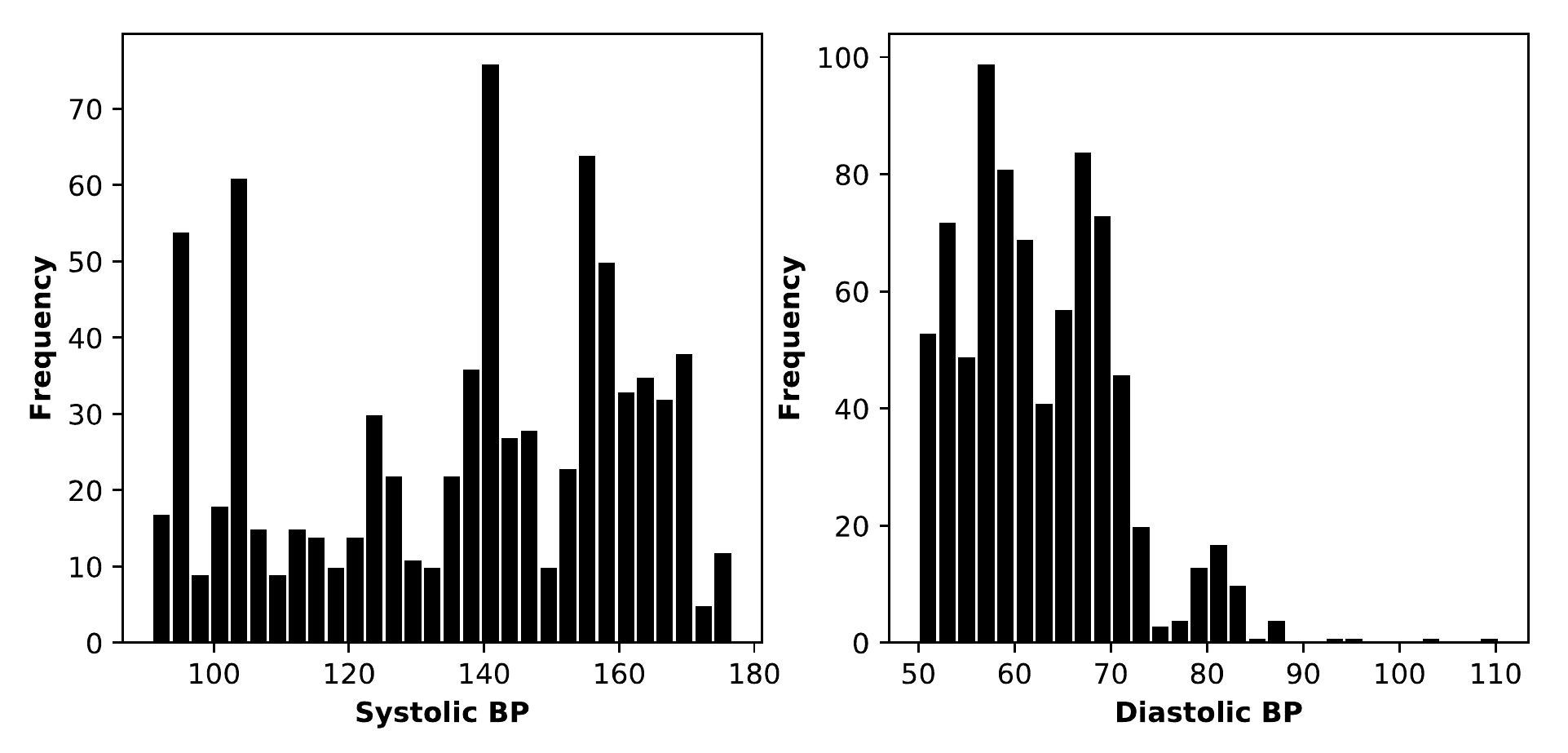}
    \caption{Distribution of Systolic BP and Diastolic BP}
        \label{fig: Hist_BP}
\end{figure}

For both datasets, we re-sampled the PPG and ABP signals at 20~Hz and re-adjusted the timings \cite{xing2016optical}. 
%(in most of the records PPG and ABP were off by around 0.25-0.35 s). 
The distribution of BP in MIMIC-II data (the portion we used in this experiment) is presented in Figure \ref{fig: Hist_BP}. After these steps, the resulting PPG signals and ABP values are ready to be used for training and evaluation of our method. 

%%%%%%%%%%%%%%%%%%%%%%%%%%%%%%%%%%%%%%%%%%%%%%%%%%%%%%%%%%%%%%%%%%%%%%%%%%%%%%%%
\section{Experimental Results} 
\label{sec: results}
%As a key metric to evaluate the proposed framework, we use the Absolute Error (AE) for each sample calculated as:
%\[
%AE_i = |BP_{T_i} - BP_{E_i}|
%\]
%in which $BP_{T_i}$ is the true or target blood pressure, and $BP_{E_i}$ is the estimated blood pressure for the $i$--th sample. Accordingly, MAE is defined as the Mean Absolute Error and SD as the Standard Deviation of the Absolute Error.
The preprocessed PPG signals and BP values from both datasets (Section \ref{sec: Data-set}) are further normalized to zero mean and unit variance before being used as the input to the network. The most important nonlinear features of the input data are extracted through a CNN network, and then these features are given to a two-layer LSTM network to do the time series analysis. 
It is important to note that unlike conventional methods, feature extraction is performed as part of the training process; the trainable weights in the CNN are optimized to extract the features that are most informative toward the estimation of the BP values. We perform \textit{leave-one-window-out} validations. Hence, one $8$s time window is kept separate, and training is done on the rest of the data. Then, the trained model is evaluated on the test sample. Also for training, we do not use three time-windows on each side of the test set, to account for the $6$s overlap between two consecutive time windows. With this strategy, the training data and test data will be completely separate from one another. We are training and testing the model for each subject independent from the others. The performance of the network is evaluated using the following three metrics:

\textbf{Absolute Error (AE):} for each test sample, AE is defined as
$AE_i = |BP_{T_i} - BP_{E_i} |$,
in which $BP_{T_i}$ is the true or target blood pressure, and $BP_{E_i}$ is the estimated blood pressure when the window number $i$ is held out as the test sample. Consequently the mean value of ${AE_i}$ and the standard deviation over the \verb|number_of_samples| iterations, and over the set of subjects are calculated and reported in the Table \ref{tab: results0}. The mean and standard-deviation of $AE$ would be a rough estimate of the error value on predictions, and the robustness and reliability of the method respectively. A model is selected as a strong predictor if it produces small $MAE$ and $SDAE$ values. The method proposed in this paper, to the best of our knowledge, outperforms other frameworks and techniques that are available until this time.

\begin{table}[bt]
    \centering
    \vspace{0.16cm}
    \caption{Performance comparison with prior works}
    \begin{tabular}{l l r r r r r }
    \specialrule{0.1em}{0.1em}{.1em} \\[-1.5ex] %\\[-1.5\medskipamount]
       \multirow{2}{*}{Work}& 
       \multirow{2}{*}{Dataset}& 
       \multicolumn{2}{l}{\textbf{SBP}} && \multicolumn{2}{l}{\textbf{DBP}} \\
       \cline{3-4} \cline{6-7} \\[-1ex]
        && MAE & SD && MAE & SD \\
        \hline \\[-1ex]
        \cite{7590775}  & MIMIC-II & 4.47 & 6.85 && 3.21 & 4.72 \\[0pt]
        \cite{article_M.Liu} & MIMIC-II & 8.54 & - && 4.34 & - \\[0pt]

%        \cite{6944640} & 32 & 0.78 & 13.1 && 0.59 & 10.23 \\[0pt]
%        \cite{lass2004continuous} & 22 & 3.22 & 8.02 && 3.13 & 4.82\\[0pt]
        \cite{MOUSAVI2019196} & MIMIC-II & 3.97 & 7.99 && 2.43 & 3.37\\[0pt]
        \cite{6555424}  & MIMIC-II & 3.80 & 3.46 && 2.21 & 2.09\\[0pt]
        \textbf{Our work}  & \textbf{MIMIC-II} & \textbf{3.70} & \textbf{3.07} && \textbf{2.02} & \textbf{1.76}\\[2pt]
        \cite{Zhang:2017:SMC:3055635.3056634} & UQVSD
        & 11.64 & 8.20 && 7.62 & 6.78\\[0pt]
        \cite{7592189} & UQVSD & 4.77& 7.68 && 3.67 & 5.69\\[0pt]
        \textbf{Our work} & \textbf{UQVSD} & \textbf{3.91} & \textbf{4.78} && \textbf{1.99} & \textbf{2.45}\\[0pt]
        \hline
    \end{tabular}
    \vspace{-4ex}
    \label{tab: results0}
\end{table}

\textbf{BHS:} As a second metric, we consider a standard commonly used for blood pressure measurements accuracy, provided by the British Hypertension Society (BHS) \cite{Brien531}, which grades the measurement accuracy into three groups (A-C). This metric measures the fraction of measurements (or estimations) which are within a certain range -- $5$, $10$ and $15$ mmHg -- of the target values for unobserved data. Based on this metric, our proposed model (to the best of our knowledge) is the only method that gets Grade \textbf{\textit{A}} for both SBP and DBP (in predictions based on PPG only). Detailed results are given in Table \ref{tab: results1}. For comparison, authors in \cite{MOUSAVI2019196} report grade \textit{A} for DBP and below \textit{C} for SBP estimation on MIMIC-II dataset. Also authors in \cite{7592189} report grade \textit{A} for DBP, and \textit{B} for SBP estimation on UQVSD dataset.

\begin{table}[b]
    \centering
    \vspace{-2ex}
    \caption{BHS Standard\cite{Brien531} vs. our results}
    \begin{subtable}[h]{.48\textwidth}
        \centering
        \caption{BHS Standard Minimum Requirements}
        \begin{tabular}{l c c c} 
         \specialrule{0.1em}{0.1em}{.1em} \\[-1.5ex] %\\[-1.5\medskipamount]
        \multirow{2}{*}{} & \multicolumn{3}{l}{\textbf{Cumulative Percentage  Error}}\\
        \cline{2-4}\\[-1ex]
        & $\leq 5$ mmHg & $\leq 10$ mmHg & $\leq 15$ mmHg\\ [.5ex]
        \hline \\[-1.5ex]
        \textbf{Grade A} & 60\% & 85\% &  95\%\\ [.5ex]
        \textbf{Grade B} & 50\% & 75\% &  90\% \\ [.5ex]
        \textbf{Grade C} & 40\% & 65\% &  85\% \\ [.5ex]
        \specialrule{0.1em}{0.1em}{.1em} \\[-1.5ex]
        \end{tabular}
        \label{tab: results11}
    \end{subtable}
    \\[3ex]
    \begin{subtable}[h]{.48\textwidth}
        \centering
        \caption{Our results on the two datasets}
        \begin{tabular}{l l c c c} 
         \specialrule{0.1em}{0.1em}{.1em} \\[-1.5ex]
        && \multicolumn{3}{l}{\textbf{Cumulative Percentage Error (Grade)}}\\
        \cline{3-5}\\[-1ex]
        {\textbf{Dataset}} & {\textbf{Signal}} & {$\leq 5$ mmHg} & {$\leq 10$ mmHg} & {$\leq 15$ mmHg}\\ [.5ex]
        \hline \\[-1.5ex]
        MIMIC-II & SBP & 77\% (\textbf{A}) & 92\% (\textbf{A})& 96\% (\textbf{A})\\
                 & DBP & 93\% (\textbf{A}) & 97\% (\textbf{A})& 99\% (\textbf{A})\\[5pt]
        UQVSD    & SBP & 75\% (\textbf{A}) & 92\% (\textbf{A})& 96\% (\textbf{A})\\
                 & DBP & 92\% (\textbf{A}) & 98\% (\textbf{A})& 99\% (\textbf{A})\\
        \specialrule{0.1em}{0.1em}{.1em} \\[-1.5ex]
        \end{tabular}
        \label{tab: results12}
    \end{subtable}
    \label{tab: results1}
    \vspace{-2ex}
\end{table}

\textbf{AAMI:} As the third metric, we consider a different standard provided by the US Association for the Advancement of Medical Instrumentation (AAMI) \cite{association2003american}. Based on this Standard, a measurement algorithm is valid if the ME (Mean Error) of the measurements is below $5$ mmHg and the SD of Errors is smaller than $8$ mmHg. The results are presented in Table \ref{tab: results2}. Our model passes the AAMI standard criteria for both SBP and DBP on both dataset. For comparison, our proposed model produces better error values (lower mean and $\sigma$) compared to \cite{7592189} on UQVSD data. On MIMIC-II dataset we get lower error $\sigma$ but higher error mean compared to \cite{MOUSAVI2019196}.

\begin{table}[bt]
    \centering
    \vspace{.15cm}
    \caption{Comparison of results with AAMI Standard \cite{association2003american}}
    \begin{tabular}{l l p{1.3cm} p{1.5cm}} 
     \specialrule{0.1em}{0.1em}{.1em} \\[-1.5ex]
    && Mean Error & $\sigma$ of Error\\
    \hline \\[-1.5ex]
    \multicolumn{2}{l}{AAMI Criteria:} & $\leq 5$ & $\leq 8$\\ [1ex]
    \hline \\[-1ex]
    MIMIC-II & SBP & 0.21 & 6.27\\
    &DBP & 0.24 & 3.40\\[1ex]
    UQVSD & SBP & 0.52 & 6.16\\
    & DBP & 0.20 & 3.15\\
    \specialrule{0.1em}{0.1em}{.1em} \\[-1.5ex]
    \end{tabular}
    %\vspace{-3ex}
   \label{tab: results2}
\end{table}

\begin{figure}[btp]
    \centering
    \includegraphics[width =.48\textwidth]{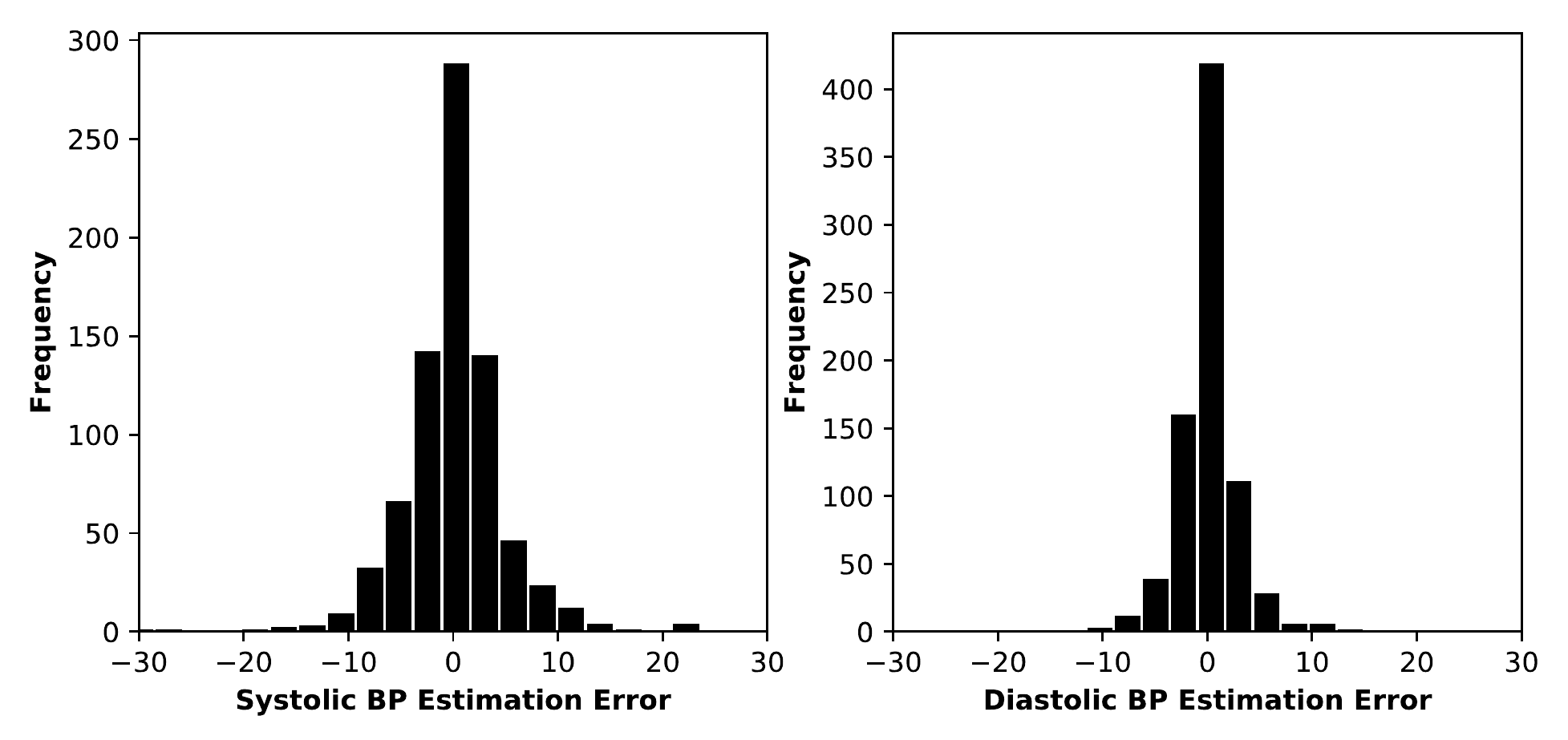}
    \caption{Prediction Error for SBP (left), and DBP (right)}
        \label{fig: Hist_BP_Error}
    \vspace{-3ex}
\end{figure}

The distribution of the prediction error for one subject from the MIMIC-II dataset is depicted in Figure \ref{fig: Hist_BP_Error}. Based on the histograms, prediction errors are mostly small. However, in some rare cases (less than 1\% of times) the estimation error for SBP goes beyond the $20$ mmHg range from the target value. For DBP, that never happened in our experiments.

\begin{figure}[btp]
     \centering
     \vspace{-1ex}
     \includegraphics[width =0.48\textwidth]{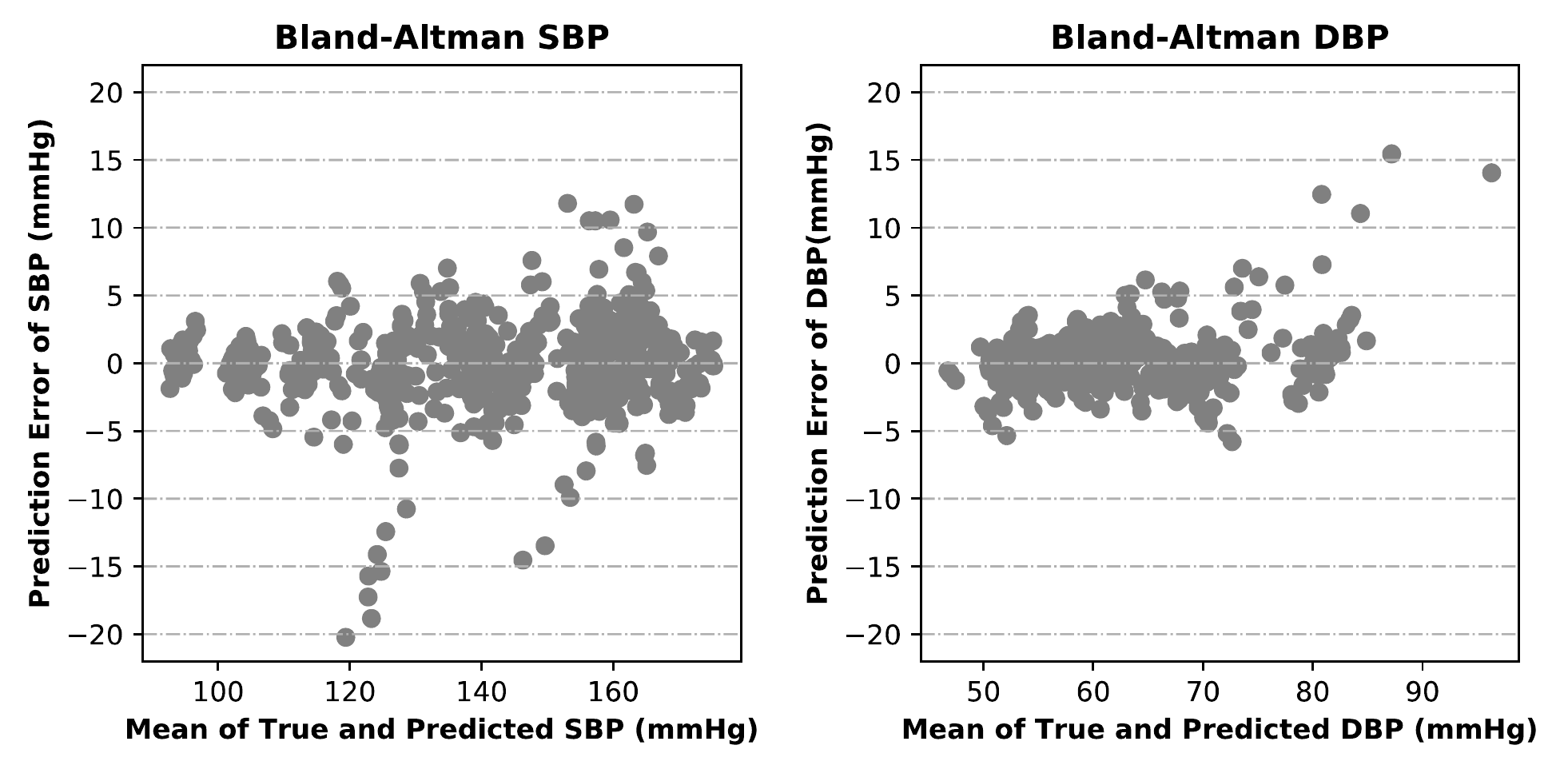}
     \caption{Bland-Altman plot for SBP and DBP}
     \label{fig: Bland_Alt}
     \vspace{-3.5ex}
\end{figure}

The Bland-Altman plot for the predictions on the same subject is presented in Figure \ref{fig: Bland_Alt}. Predictions on lower values of BP -- left side of each plot - are more reliable (smaller error). Note that these samples correspond to when the subject is likely to be less active (potentially at rest), at which the PPG signal is expected to be less noisy and more reliable.

% \subsection*{Complexity Analysis}

% The table below summarizes the complexity of the proposed framework.

% \begin{table}
%     \centering
%     \caption{\small Complexity Analysis}
%     \begin{tabular}{ r | l l} 
%      \specialrule{0.1em}{0.1em}{.1em} \\[-1.5ex]
%     & {\small Hyper-parameters} & {\small Trainable Parameters}\\
%     \hline \\[-1.5ex]
%     CNN & M & N\\ [.5ex]
%     LSTM1 & M & N\\
%     LSTM2 & M & N\\
%     Dense & M & N\\
%     \specialrule{0.1em}{0.1em}{.1em} \\[-1.5ex]
%     \end{tabular}
%   \label{tab: complexity}
% \end{table}

%%%%%%%%%%%%%%%%%%%%%%%%%%%%%%%%%%%%%%%%%%%%%%%%%%%%%%%%%%%%%%%%%%%%%%%%%%%%%%%%
\section{Conclusions and future work} \label{sec: Conclusions}
Hypertension or high blood pressure is closely related to cardiovascular disease. While continuous and portable BP monitoring is of high importance, conventional cuff-based BP measurement devices are primarily confined to clinical settings. The method proposed in this paper is a cuff-less, noninvasive and feasible method which estimates Blood Pressure with high accuracy based on only single channel PPG signal. PPG sensors are convenient to use, simple and low cost, and are integrated in most of newer wearable devices. We use Convolutional Neural Network to extract the optimal set of features from the input, and then use a LSTM network to capture the temporal correlations in the extracted features (similar to \cite{8607019}). Results of the proposed method outperform previous works that only use PPG signal as input. Importantly, the proposed method ranks as \textbf{\textit{A}} in the BHS Standard for both SBP and DBP, complies with the AAMI Standard, and outperforms all other similar works in terms of prediction error (as well as BHS criteria).

Importantly, the proposed method is feasible for practical applications in commercial products. The processing power required to execute the model (and more importantly, to keep it updated with the new incoming data from each subject) might be too intensive to be executed on the edge layer in an IoT architecture. However, the raw PPG data can be transferred to the cloud, and the prediction output can be sent back to the user through a smart-watch or smart-phone, all in a real-time scenario. The development of such architecture is left to future work.
\vspace{-2ex}
\bibliographystyle{ieeetr}
\bibliography{main}
%bibliographystyle{unsrt} %ieeetr IEEEtran  unsrt
%\vspace{12pt}

\end{document}